\documentclass[sigconf]{acmart}

\usepackage{booktabs} 
\usepackage{caption}
\usepackage{subcaption}
\usepackage{footmisc}
\usepackage{balance}

\setcopyright{rightsretained}

\renewcommand\footnotetextcopyrightpermission[1]{} 
\begin{document}
\title{Deep Learning based Large Scale Visual Recommendation and Search for E-Commerce}

\author{Devashish Shankar, Sujay Narumanchi, Ananya H A, \\ Pramod Kompalli, Krishnendu Chaudhury}
\affiliation{%
  \institution{Flipkart Internet Pvt. Ltd.,\\ Bengaluru, India.}
}
\email{{devashish.shankar, sujay.nv, ananya.h, pramod.kompalli, krishnendu}@flipkart.com}

\renewcommand{\shortauthors}{Shankar et. al.}

\begin{abstract}
In this paper, we present a unified end-to-end approach to build a large scale Visual Search and Recommendation system for e-commerce. Previous works have targeted these problems in isolation. We believe a more effective and elegant solution could be obtained by tackling them together. We propose a unified Deep Convolutional Neural Network architecture, called VisNet \footnote{Our training code is open-sourced at \href{https://github.com/flipkart-incubator/fk-visual-search}{https://github.com/flipkart-incubator/fk-visual-search}.}, to learn embeddings to capture the notion of visual similarity, across several semantic granularities. We demonstrate the superiority of our approach for the task of image retrieval, by comparing against the state-of-the-art on the \textit{Exact Street2Shop} \cite{WhereToBuy} dataset. We then share the design decisions and trade-offs made while deploying the model to power Visual Recommendations across a catalog of 50M products, supporting 2K queries a second at Flipkart, India's largest e-commerce company. The deployment of our solution has yielded a significant business impact, as measured by the conversion-rate.
\end{abstract}

\keywords{Deep Learning, Computer Vision, Visual Search, Image Retrieval, Distributed Systems, Recommender Systems, E-Commerce}

\maketitle

\section{Introduction}
\label{sec:intro}
A large portion of sales in the e-commerce domain is driven by fashion and lifestyle, which constitutes apparel, footwear, bags, accessories, etc. Consequentially, a rich user discovery experience for fashion products is an important area of focus. A differentiating characteristic of the fashion category is that a user's buying decision is primarily influenced by the product's visual appearance. Hence, it is crucial that ``Search" and ``Recommendations", the primary means of discovery, factor in the visual features present in the images associated with these items \cite{PinterestVisualSearch, Etsy, VisualSimilarityProductDesign}. 

Traditional e-commerce search engines are lacking in this regard since they support only text-based searches that make use of textual metadata of products such as attributes and descriptions. These methods are less effective, especially for the fashion category, since detailed text descriptions of the visual content of products are difficult to create or unavailable. 

This brings us to the problem of image-based search or Visual Search where the goal is to enable users to search for products by uploading a picture of an item they like - this could be a dress worn by a celebrity in a magazine or a nice looking handbag owned by a friend. While it is important to find an exact match of the query item from the catalog, retrieving a ranked list of similar items is still very significant from a business point of view.

A related problem is that of image-based recommendations or Visual Recommendations. A user interested in buying a particular item from the catalog may want to browse through visually similar items before finalising the purchase. These could be items with similar colors, patterns and shapes. Traditional recommender systems \cite{AmazonCF, BPRMF} that are based on collaborative filtering techniques fail to capture such details since they rely only on user click / purchase activity and completely ignore the image content. Further, they suffer from the `cold start' problem - newly introduced products do not have sufficient user activity data for meaningful recommendations.

\begin{figure}[t]
\centering
\begin{subfigure}[t]{4cm}
\centering
\includegraphics[width=1.75cm]{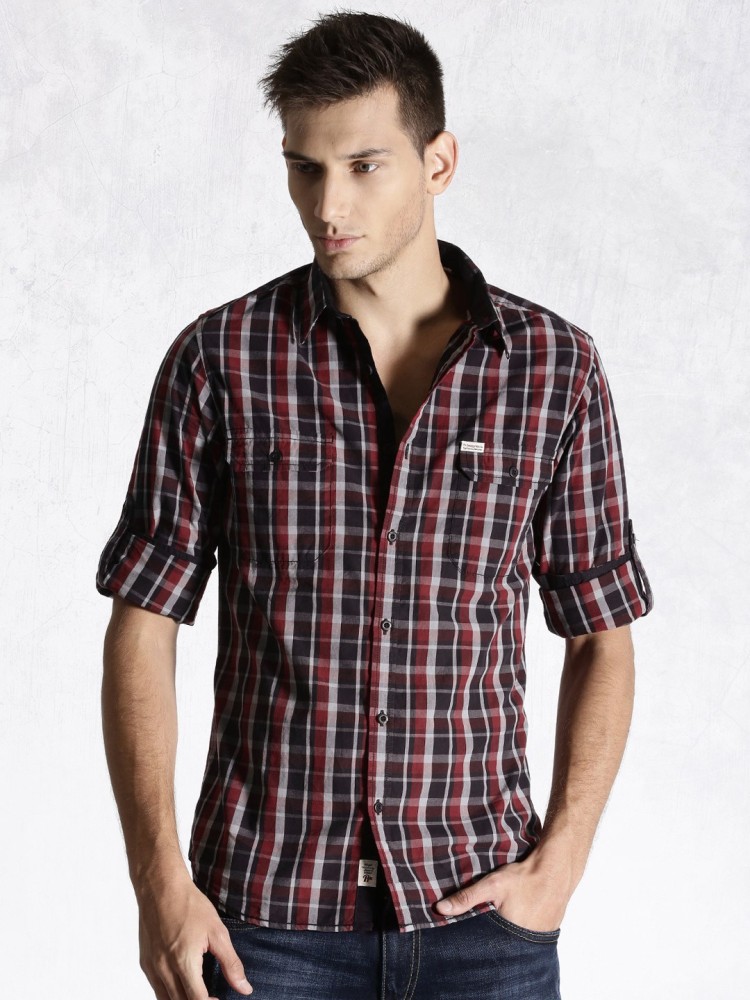}
\includegraphics[width=1.75cm]{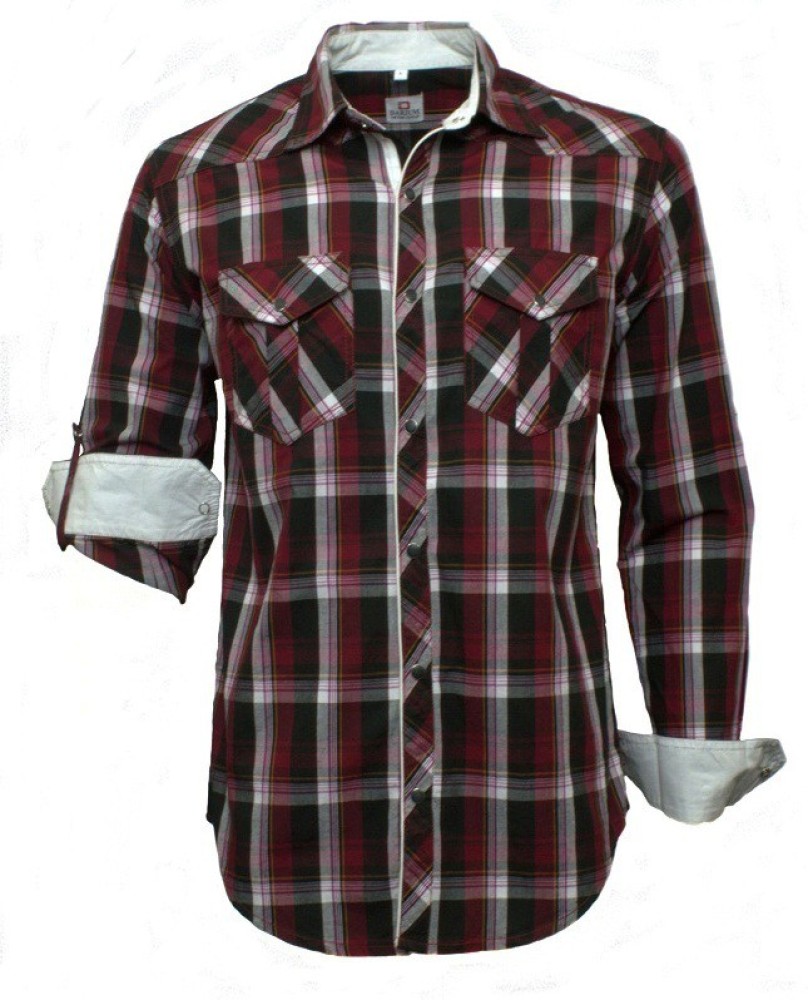}
\subcaption{Similar catalog items with and without human model}
\label{CIVR_match_shirt_wwo_model}
\end{subfigure} 
\hfill
\begin{subfigure}[t]{4cm}
\centering
\includegraphics[width=1.75cm]{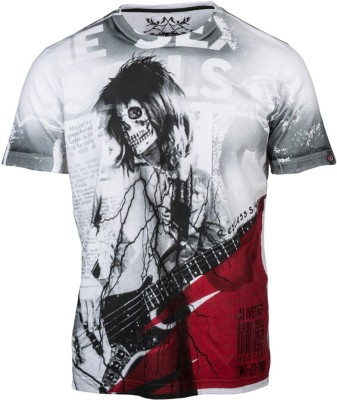}
\includegraphics[width=1.75cm]{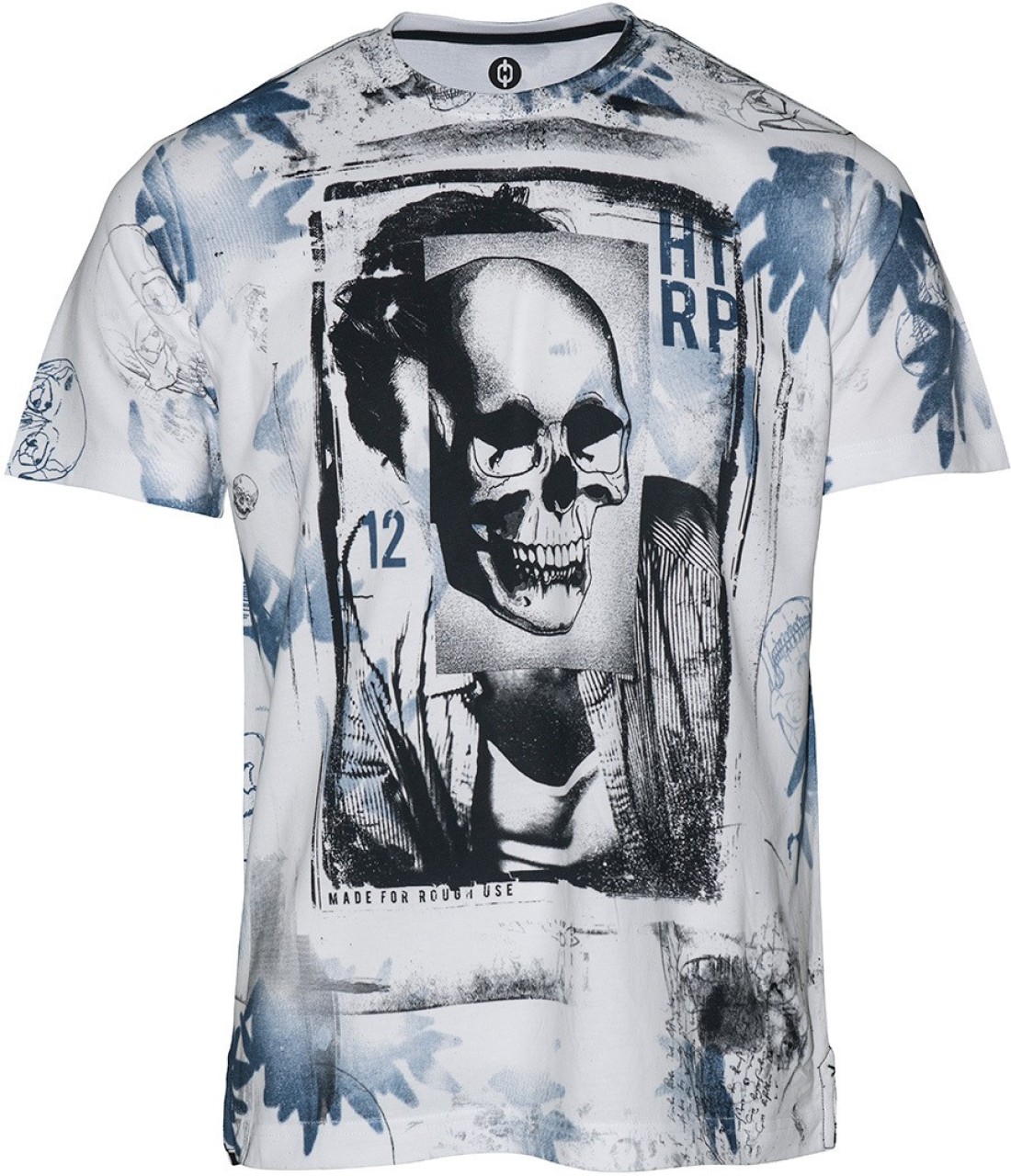}
\subcaption{Concept based similarity \mbox{across} spooky printed t-shirts}
\label{CIVR_match_concept}
\end{subfigure}
\vspace{0.3cm}\\
\begin{subfigure}[t]{4.1cm}
\centering
\includegraphics[width=1.3cm]{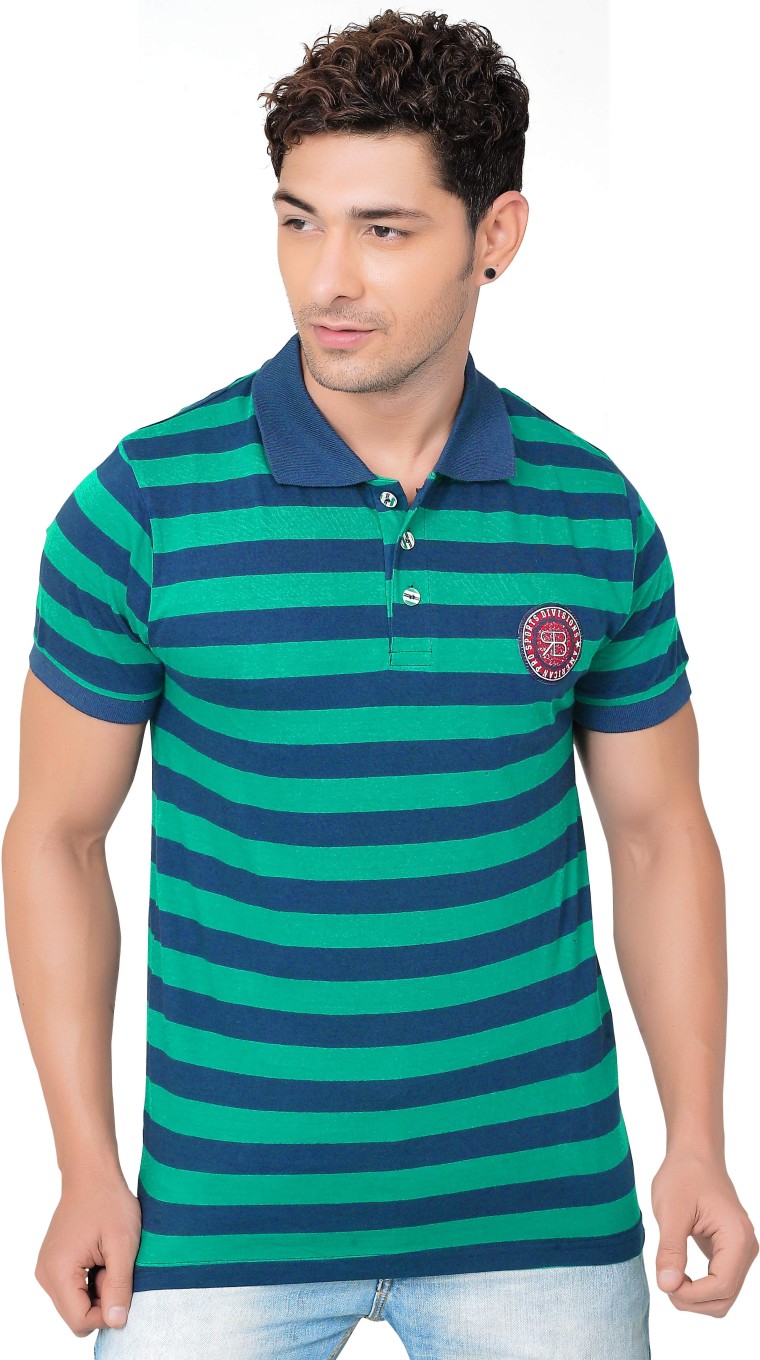}
\hspace{0.1cm}
\includegraphics[width=1.35cm]{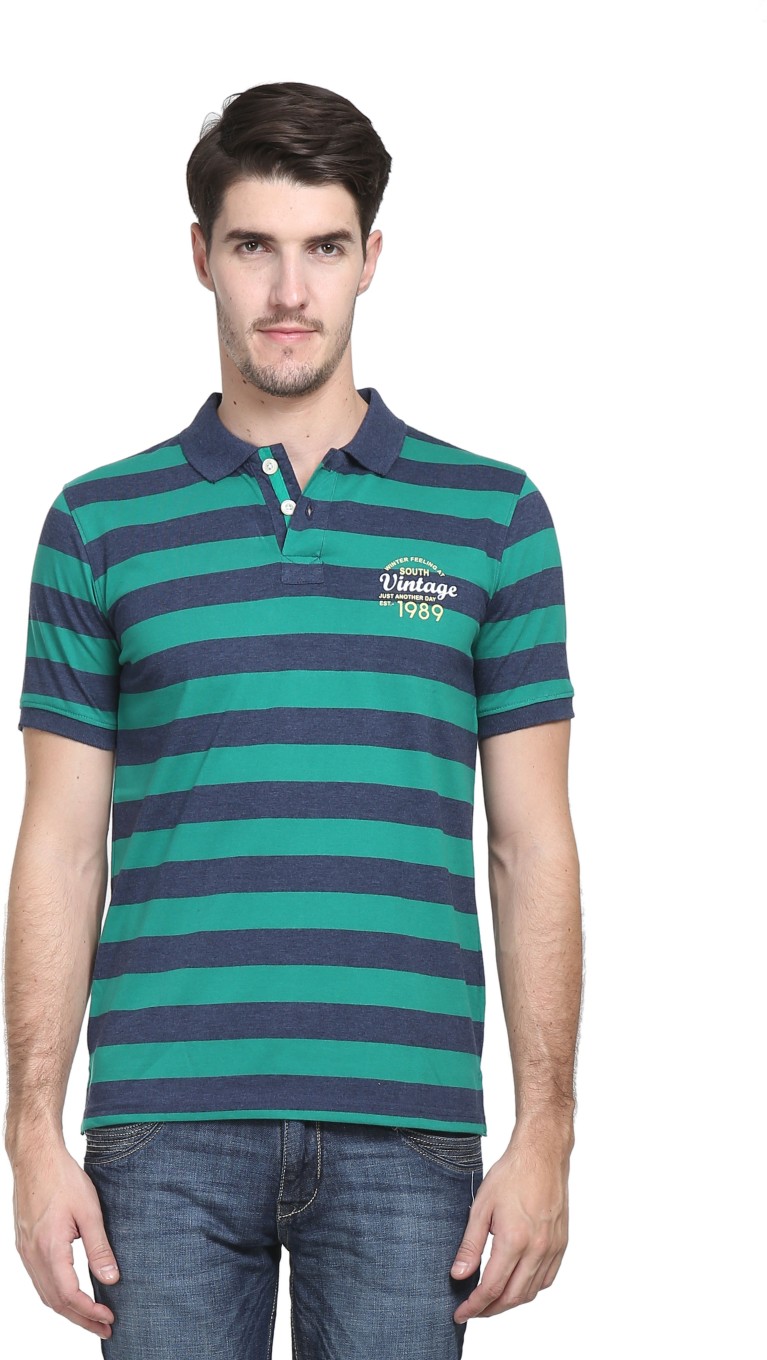}
\includegraphics[width=1.1cm]{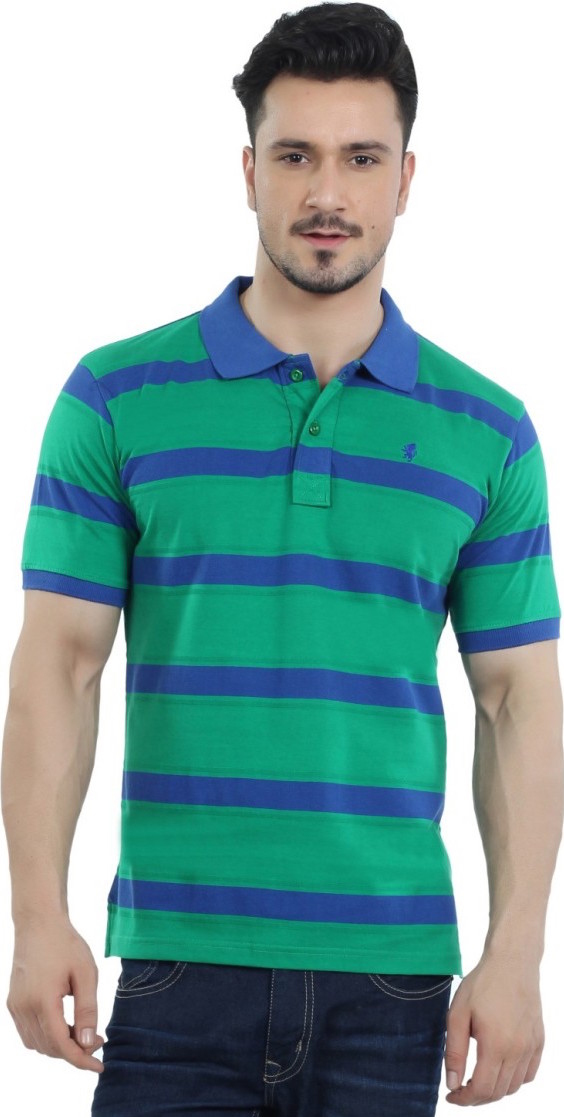}
\subcaption{Detail based similarity via spacing and thickness of stripes}
\label{CIVR_match_details}
\end{subfigure}
\hfill
\begin{subfigure}[t]{4cm}
\centering
\includegraphics[width=1.75cm]{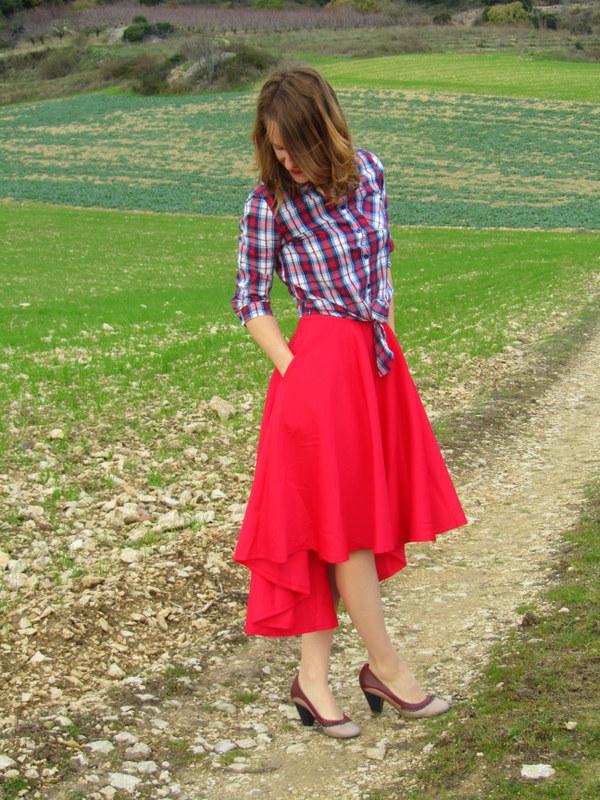}
\includegraphics[width=1.75cm]{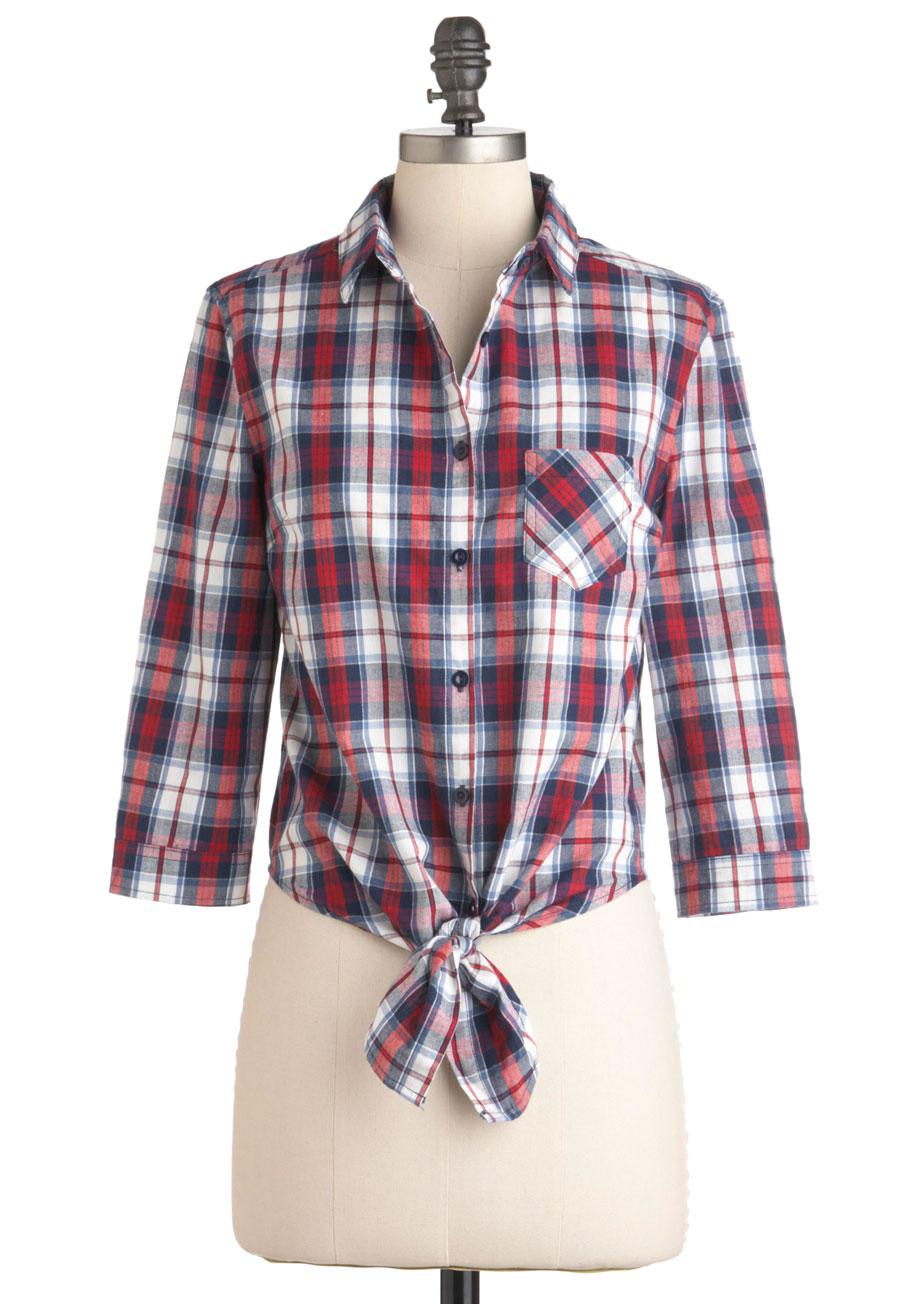}
\subcaption{Wild Image similarity across radically different poses}
\label{WIVR_lookup}
\end{subfigure}
\caption{Visual Similarity Challenges}
\vspace{-0.6cm}
\end{figure}


In this paper, we address the problems of both Visual Recommendations (retrieving a ranked list of catalog images similar to another catalog image) and Visual Search (retrieving a ranked list of catalog images similar to a ``wild" (user-uploaded) image). The core task, common to both, is quantitative estimation of visual similarity between two images containing fashion items. This is fraught with several challenges, as outlined below. We start with the challenges in visual recommendation, where we deal only with catalog images. These images are taken under good lighting conditions, with relatively uniform backgrounds and the foreground is typically dominated by the object of interest. Nevertheless, there can be tremendous variation between images of the same item. For instance, image of a dress item laid flat on a surface is deemed very similar to the image of the same shirt worn by a human model or mannequin (see Figure~\ref{CIVR_match_shirt_wwo_model}). Thus, the similarity engine must be robust to the presence or absence of human model or mannequins, as well as to different looks, complexions, hair colors and poses of these models / mannequins. Furthermore, the human notion of item similarity itself can be quite abstract and complex. For instance, in Figure~\ref{CIVR_match_concept}, the two t-shirts are deemed similar by most humans because they depict a spooky print. Thus, the similarity engine needs to have some understanding of higher levels of abstractions than simple colors and shapes. That does not mean the engine can ignore low level details. Essentially, the engine must pay attention to image characteristics at multiple levels of abstraction. For instance, t-shirts should match t-shirts, round-necked t-shirts should match round-necked t-shirts, horizontal stripes should match horizontal stripes and closely spaced horizontal stripes should match closely spaced horizontal stripes. Figure~\ref{CIVR_match_details} is an example where low-level details matter, t-shirts in the left and center are more similar compared to the left and right due to similar spacing between the stripes.

In addition to the above challenges, visual search brings its own set of difficulties. Here the system is required to estimate similarities between user uploaded photos (``wild images") and catalog images. Consequently, the similarity engine must deal with arbitrarily complex backgrounds, extreme perspective variations, partial views and poor lighting. See Figure~\ref{WIVR_lookup} for an example.

The main contribution of this paper is an end-to-end solution for large scale Visual Recommendations and Search. We share the details of our model architecture, training data generation pipeline as well as the architecture of our deployed system. Our model, \textit{VisNet}, is a Convolutional Neural Network (CNN) trained using the triplet based deep ranking paradigm proposed in \cite{TripletPaper}. It contains a deep CNN modelled after the VGG-16 network \cite{vgg}, coupled with parallel shallow convolution layers in order to capture both high-level and low-level image details simultaneously (see Section \ref{ssec:cnnarch}). Through extensive experiments on the \textit{Exact Street2Shop} dataset created by Kiapour et al. in \cite{WhereToBuy}, we demonstrate the superiority of VisNet over previous state-of-the-art.

We  also present a semi-automatic training data generation methodology that is critical to training VisNet. Initially, the network is trained only on catalog images. Candidate training data is generated programmatically with a set of Basic Image Similarity Scorers from which final training data is selected via human vetting. This network is further fine-tuned for the task of Visual Search, with wild image training data from the Exact Street2Shop dataset. Details of our training data generation methodology can be found in Section \ref{ssec:TD}.

Additionally, we present the system that was designed and successfully deployed at Flipkart, India's biggest e-commerce company with over 100 million users making several thousand visits per second. The sheer scale at which the system is required to operate presents several deployment challenges. The fashion catalog has over 50 million items with 100K additions / deletions per hour. Consequently, keeping the recommendation index fresh is a challenge in itself. We provide several key insights and the various trade-offs involved in the process of building a horizontally scalable, stable, cost-effective, state of the art Visual Recommendation engine, running on commodity hardware resources.

%
\section{Related Work}
\label{sec:RelatedWork}
\begin{figure*}[t]
\centering
\framebox{\includegraphics[trim=0cm 5cm 0cm 5cm, clip=true,width=16.0cm]{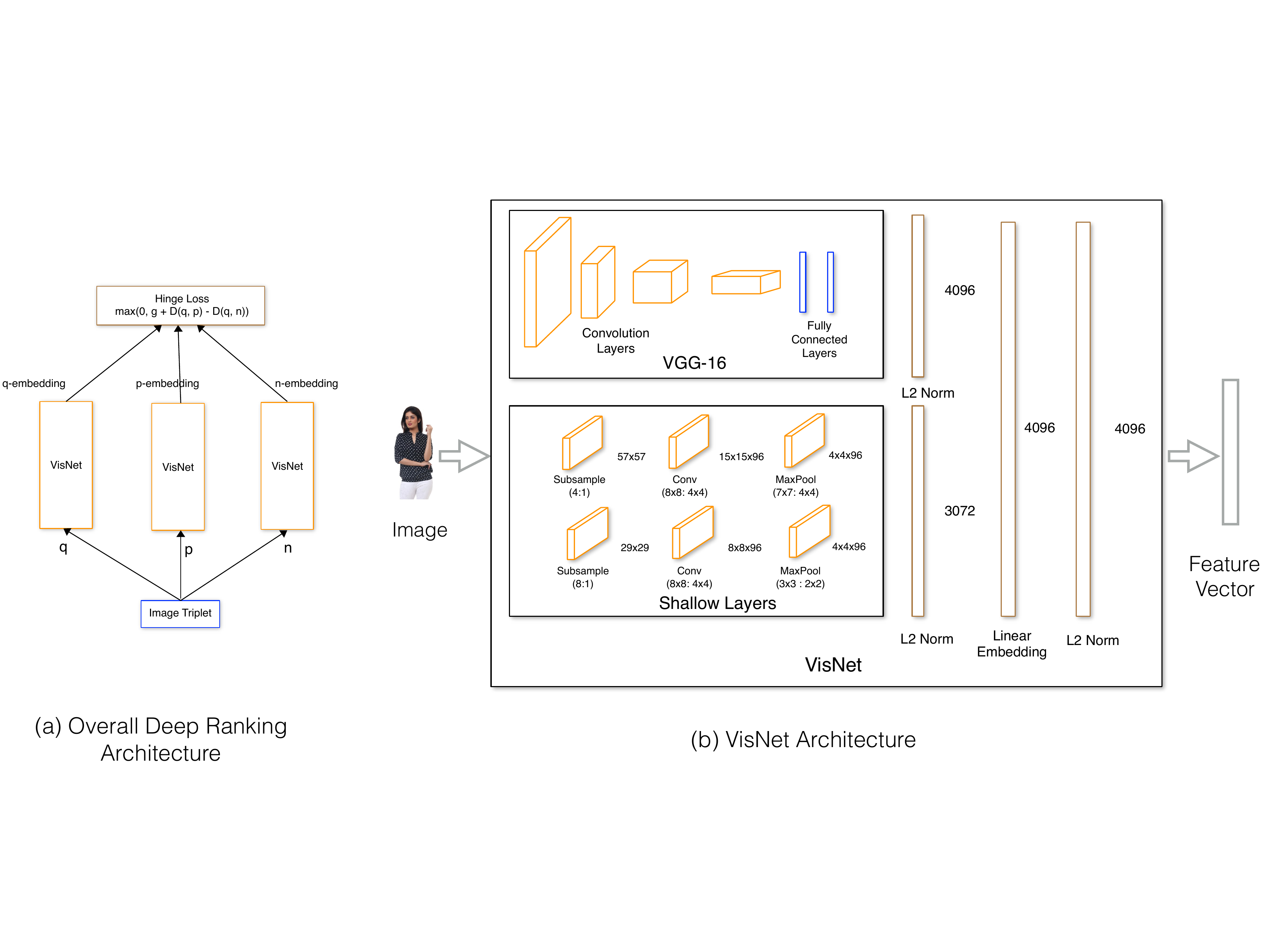}}
\caption{Deep Ranking CNN Architecture}
\label{fig:CNNarch}
\end{figure*}

\textbf{Image Similarity Models:}
Image similarity through learnt models on top of traditional computer vision features like SIFT \cite{SIFT}, HOG \cite{HOG} have been studied in \cite{midlevel,OASIS,learninginvariance}. However, these models are limited by the expressive power of these features. In recent years, Deep CNNs have been used with unprecedented success for object recognition \cite{AlexNet,GoogLeNet,vgg}. As is well known, successive CNN layers represent the image with increasing levels of abstraction. The final layers contain abstract descriptor vectors for the image, robust to variations in scale, location within image, viewpoint differences, occlusion etc. However, these descriptors are not very good at visual similarity estimation since similarity is a function of both abstract high level concepts (shirts match other shirts, not shoes) as well as low level details (pink shirts match pink shirts, striped shirts match striped shirts, closely spaced stripes match closely spaced stripes more than widely spaced stripes etc). The latter is exactly what the object detection network has learnt to ignore (since it wants to recognise a shirt irrespective of whether it is pink or yellow, striped or chequered). Stated in other words, the object recognition network focusses on features common to all the objects in that category, ignoring fine details that are often important for similarity estimation. This reduces its effectiveness as similarity estimator. We share results of using features from the FC7 layer of the AlexNet \cite{AlexNet} in Section \ref{sec:Results}.

Siamese networks \cite{SiameseNetworks} with contrastive loss can also be used for similarity assessment. Here, the network comprises of a pair of CNNs with shared weights. It takes a pair of images, as input. The ground truth labels the image pair as similar or dissimilar. The advantage of the Siamese network is that it directly trains for the very problem (of similarity) that we are trying to solve - as opposed to, say, training for object detection and then using the network for similarity estimation. However, as it is trained to make binary decisions (yes / no) on whether two images are similar, it fails to capture fine-grained similarity. Furthermore, a binary decision on whether two images are similar is very subjective. Consequently, ground truth tends to be contradictory and non-uniform - replication (where multiple human operators are given the same ground truth and their responses are aggregated) mitigates the problem but does not quite solve it.

Image similarity using Triplet Networks has been studied in \cite{TripletPaper,featurelearninghashcoding}. The pairwise ranking paradigm used here is essential for learning fine-grained similarity. Further, the ground truth is more reliable as it is easier to label for relative similarity than absolute similarity.

\textbf{Image Retrieval:}
Image Retrieval for fashion has been extensively studied in \cite{RobustSiameseVisualSearch, PinterestVisualSearch, DeepFashion, WhereToBuy}. Kiapour et al. \cite{WhereToBuy} focus on the Exact Street to Shop problem in order to find the same item in online shopping sites. They compare a number of approaches. Their best performing model is a two-layer neural network trained on final layer deep CNN features to predict if two items match. They have also curated, human labelled and made available, the Exact Street2Shop dataset. The focus of their paper is \emph{exact street to shop} (retrieving the exact match of a query object in a wild/street image from catalog/shop). Our focus is on \emph{visual recommendations and search}, which includes exact retrieval of query item as well as retrieval of similar items and as such is a superset of the problem solved by them.

Wang et al. \cite{RobustSiameseVisualSearch} propose a deep Siamese Network with a modified contrastive loss and a multi-task network fine tuning scheme. Their model beats the previous state-of-the-art by a significant margin on the Exact Street2Shop dataset. However, their model being a Siamese Network suffers from the same limitations discussed above. Liu et al. \cite{DeepFashion} propose ``FashionNet" which learns clothing features by jointly predicting clothing attributes and landmarks. They apply this network to the task of street-to-shop on their own DeepFashion dataset. The primary disadvantage of this method is the additional training data requirement of landmarks and clothing attributes. The street-to-shop problem has also been studied in \cite{cross_scenario}, where a parts alignment based approach was adopted.

A related problem, clothing parsing, where image pixels are labelled with a semantic apparel label, has been studied in \cite{whoblockswho,paperdoll,parsingclothing,parselet,weakcolor}. Attribute based approach to clothing classification and retrieval has been studied in \cite{apparelclassification,descclothing,stylefinder}. Fine-tuning of pre-trained deep CNNs have  been used for clothing style classification in \cite{deeplearningforclothing}. In \cite{rapidclothing}, a mid level visual representation is obtained from a pre-trained CNN, on top of which a latent layer is added to obtain a hash-like representation to be used in retrieval.

There has also been a growing interest in developing image based retrieval systems in the industry. For instance, Jing et al. \cite{PinterestVisualSearch} present a scalable and cost effective commercial Visual Search system that was deployed at Pinterest  using widely available open source tools. However, their image descriptor, which is a combination of pre-trained deep CNN features and salient color signatures, is limited in its expressive power. Lynch et al. \cite{Etsy} use multimodal (image + text) embeddings to improve the quality of item retrieval at Etsy. However, the focus of this paper is only on images.

\section{Overall Approach}
\label{sec:OverallApproach}

\subsection{Deep Ranking CNN Architecture}
\label{ssec:cnnarch}Our core approach consists of training a Convolutional Neural Network (CNN) to generate embeddings that capture the notion of visual similarity. These embeddings serve as visual descriptors, capturing a complex combination of colors and patterns. We use a triplet based approach with a ranking loss to learn embeddings such that the Euclidean distance between embeddings of two images measures the (dis)similarity between the images. Similar images can be then found by k-Nearest-Neighbor searches in the embedding space.


Our CNN is modelled after \cite{TripletPaper}. We replace AlexNet with the 16-layered VGG network  \cite{vgg} in our implementation. This significantly improves our recall numbers (see section~\ref{sec:Results}). Our conjecture is that the dense (stride $1$) convolution with small receptive field ($3\times3$) digests pattern details much better than the sparser AlexNet. These details may be even more important in the product similarity problem than the object recognition problem.

Each training data element is a triplet of $3$ images, $<q, p, n>$, a query image ($q$), a positive image ($p$) and a negative image ($n$). It is expected that the pair of images $\left(q, p\right)$ are more visually similar compared to the pair of images $\left(q, n\right)$. Using triplets enables us to train a network to directly rank images instead of optimising for binary/discriminatory decisions (as done in Siamese networks). It should be noted that here the training data needs to be labeled only for \emph{relative} similarity. This is much more well defined than \emph{absolute} similarity required by Siamese networks. On the one hand, this enables us to generate a significant portion of the training data programmatically, there is no need to find an absolute match / no-match threshold for the candidate training data generator programs that we use. On the other hand, this also allows the human operators that verify those candidate triplets to be much faster and consistent.

\begin{figure}
\centering
\includegraphics[trim=0.5cm 0cm 0cm 0cm, clip=true,width=8.0cm]{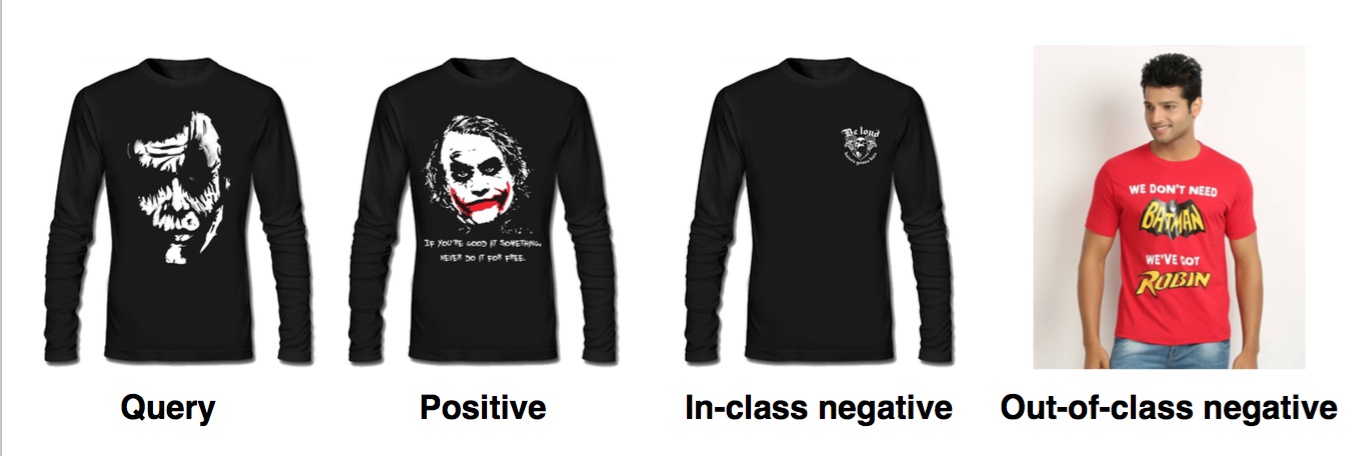}	
\vspace{-0.2cm}
\caption{An example Triplet used for training}
\vspace{-0.4cm}
\label{fig:TripletExample}
\end{figure}

We use two different types of triplets, in-class triplets and out-of-class triplets (see Figure \ref{fig:TripletExample} for an example). The in-class triplets help in teaching the network to pay attention to nuanced differences, like the thickness of stripes, using hard negatives. These are images that can be deemed similar to the query image in a broad sense but are less similar to the query image when compared to the positive image due to fine-grained distinctions. This enables the network to learn robust embeddings that are sensitive to subtle distinctions in colors and patterns. The out-of-class triplets contain easy negatives and help in teaching the network to make coarse-grained distinctions.

During training, the images in the triplet are fed to  $3$ sub-networks with shared weights (see Figure~\ref{fig:CNNarch}a).  Each sub-network generates an embedding or feature vector, thus $3$ embedding vectors, $\vec{q}$, $\vec{p}$ and $\vec{n}$ are generated - and fed into a \emph{Hinge Loss} function 
\begin{equation}
L = max \left( 0, g + D\left( \vec{q}, \vec{p}\right) - D\left( \vec{q}, \vec{n}\right) \right)
\end{equation}
where $D\left( \vec{x}, \vec{y}\right)$ denotes the Euclidean Distance between $\vec{x}$ and $\vec{y}$.
The hinge loss function is such that minimizing it is equivalent, in the embedding space, to pulling the points $\vec{q}$ and $\vec{p}$ closer and pushing the points $\vec{q}$ and $\vec{n}$ farther.
To see this, assume $g = 0$ for a moment. One can verify the following
\begin{itemize}
\item{Case 1. Network behaved correctly:  $D\left( \vec{q}, \vec{p} \right) - D\left( \vec{q}, \vec{n} \right) < 0 \implies loss = 0$}
\item{Case 2. Network behaved incorrectly:   $D\left(\vec{q}, \vec{p}\right) - D\left(\vec{q}, \vec{n}\right) > 0 \implies loss > 0$}
\item{More the network deviates from correct behavior, higher the hinge loss}
\end{itemize}
Introduction of a non-zero $g$ simply pushes the positive and negative images further apart, $D\left( \vec{q}, \vec{n} \right) - D\left( \vec{q}, \vec{p} \right)$ has to be greater than $g$ to qualify for no-loss.

Figure~\ref{fig:CNNarch}b shows the content of each sub-network. It has the following parallel paths
\begin{itemize}
\item{\textbf{16-Layer VGG net without the final loss layer}: the output of the last layer of this net captures abstract, high level characteristics of the input image}
\item{\textbf{Shallow Conv Layers 1 and 2}:  capture fine-grained details of the input image.}
\end{itemize}
This parallel combination of deep and shallow network is essential for the network to capture both the high level and low level details needed for visual similarity estimation, leading to better results.

During inference, any one of the sub-networks  (they are all the same, since the weights are shared) takes an image as input and generates an embedding. Finding similar items then boils down to the task of nearest neighbor search in the embedding space.

We grouped the set of product items into related categories, e.g., clothing (which includes shirts, t-shirts, tops, etc), footwear, and trained a \emph{separate} deep ranking NN for each category. Our deep networks  were implemented on Caffe \cite{caffe}.
\subsection{Training Data Generation}
\label{ssec:TD}
The training data comprises of two classes of triplets - (1) Catalog Image Triplets and (2) Wild Image Triplets \\

 \textbf{Catalog Image Triplets}: Here, the query image ($q$), positive image ($p$) and negative image ($n$) are all catalog images. During training, in order to generate  a triplet $< q, p, n >$, $q$ is randomly sampled from the set of catalog images. Candidate positive images are programmatically selected in a \emph{bootstrapping} manner by a set of basic image similarity scoring techniques, described below. It should be noted that a ``Basic Image Similarity Scorer" (BISS) need not be highly accurate. It need not have great recall (get most images highly similar to query) nor great precision (get nothing but highly similar images). The principal expectation from the BISS set is that, between all of them, they more or less identify all the images that are reasonably similar to the query image. As such, each BISS could focus on a sub-aspect of similarity, e.g., one BISS can focus on color, another on pattern etc. 
Each BISS  programmatically identifies the $1000$ nearest neighbors to $q$ from the catalog. The union of top $200$ neighbors from all the BISSs form the sample space for $p$.

Negative images, $n$, are sampled from two sets, respectively, in-class and out-of-class. In-class negative images refer to the set of $n$'s that are \emph{not} very different from $q$. They are \emph{subtly} different, thereby teaching \emph{fine grained similarity} to the network (see \cite{TripletPaper} for details). Out-of-class negative images refer to the set of $n$'s that are very  different from $q$. They teach the network to do coarse distinctions. In our case, the in-class $n$'s were sampled from the union, over all the BISSs, of the nearest neighbors ranked between $500$ and $1000$ . The out-of-class $n$'s were sampled from the rest of the universe (but within the same product category group). Our sampling was biased to contain $30\%$ in-class and $70\%$ out-of-class negatives.

The final set of triplets were manually inspected and wrongly ordered triplets were corrected. This step is important as it prevents the mistakes of individual BISSs from getting into the final machine.

Some of the basic image similarity scorers used by us are: 1) $AlexNet$, where the activations of the FC7 layer of the pre-trained AlexNet \cite{AlexNet} are used for representing the image. 2) $ColorHist$, which is the LAB color histogram of the image foreground. Here, the foreground is segmented out using Geodesic Distance Transform \cite{geodesic1,geodesic_ziss} followed by Skin Removal. 3) $PatternNet$, which is an AlexNet trained to recognize patterns (striped, checkered, solids) associated with catalog items. The ground truth is automatically obtained from catalog metadata and the FC7 embedding of this network is used as a BISS for training data generation. Note that the structured metadata is too coarse-grained to yield a good similarity detector by itself (patterns are often too nuanced and detailed to be described in one or two words, e.g., a very large variety of t-shirt and shirt patterns are described by the word printed). In section \ref{sec:Results} we provide the overall performance of each BISS, vis-a-vis the final method.\\

\textbf{Wild Image Triplets}:
VisNet was initially trained only on Catalog Image Triplets. While this network served as a great visual recommender for catalog items, it performed poorly on wild images due to the absence of complex backgrounds, large perspective variations and partial views in training data. To train the network to handle these complications, we made use of the Exact Street2Shop dataset created by Kiapour et al. in \cite{WhereToBuy}, which contains wild images (street-photos), catalog images (shop-photos) and exact street-to-shop pairs (established manually). It also contains bounding boxes around the object of interest within the wild image. Triplets were generated from the clothing subset of this dataset for training and evaluation. The query image, $q$ of the triplet was sampled from the cropped wild images (using bounding box metadata). Positive element of the triplet was always the ground truth match from catalog. Negative elements were sampled from other catalog items. In-class and out-of-class separation was done in a bootstrapping fashion, on the basis of similarity scores provided by the VisNet trained only on Catalog lmage Triplets. The same in-class to out-of-class ratio ($30\%$ to $70\%$) is used as in case of Catalog Image Triplets.\\
\\

\subsection{Object Localisation for Visual Search}
\label{ssec:localisationVS}
We observed that training VisNet on cropped wild images, i.e. only the region of interest, as opposed to the entire image yielded best results. However, this mandates cropped versions of user-uploaded images to be used during evaluation and deployment. One possible solution is to allow the user to draw the rectangle around the region of interest while uploading the image. This however adds an extra burden on the user which becomes a pain point in the workflow. In order to alleviate this issue, and to provide a seamless user experience, we train an object detection and localisation network using the Faster R-CNN (FRCNN) approach outlined in \cite{fasterrcnn}. The network automatically detects and provide bounding boxes around objects of interest in wild images, which can then be directly fed into VisNet. 

We create a richly annotated dataset by labelling several thousand images, consisting of wild images obtained from the Fashionista dataset \cite{fashionista} and Flipkart catalog images. The Fashionista dataset was chosen because it contains a large number of unannotated wild images with multiple items of interest. We train a VGG-16 FRCNN on this dataset using alternating optimisation using default hyperparameters. This model attains an average mAP of 68.2\% on the test set. In Section \ref{sec:Results}, we demonstrate that the performance of the end-to-end system, with the object localisation network to provide the region of interest, is comparable to the system with user specified bounding boxes.

\section{Implementation Details and Results}
\label{sec:Results}
\subsection{Training Details}
\label{ssec:trainingDetails}
For highest quality, we grouped our products into similar categories (e.g., dresses, t-shirts, shirts, tops,  etc. in one category called clothing, footwear in another etc.) and trained separate networks for each category. In this paper, we are mostly reporting  the clothing network numbers. All networks had the same architecture, shown in Figure~\ref{fig:CNNarch}. The clothing network was trained on 1.5 million Catalog Image Triplets and 1.5 million Wild Image Triplets. Catalog Image Triplets were generated from 250K t-shirts, 150K shirts, 30K tops, altogether about 500K dress items. Wild Image Triplets were generated from the Exact Street2Shop (\cite{WhereToBuy}) dataset which contains around 170K dresses, 68K tops and 35K outerwear items.

The CNN architecture was implemented in \textit{Caffe} \cite{caffe}. Training was done on an nVidia machine with 64GB RAM, 2 CPUs with 12 core each and 4 GTX Titan X GPUs with 3072 cores and 12 GB RAM per GPU. Each epoch (13077 iterations) took roughly 20 hours to complete. Replacing the VGG-16 network with that of AlexNet gave a 3x speedup (6 hours per epoch), but led to a significant drop in quality of results.


\subsection{Datasets and Evaluation}

\begin{figure}[t]
\centering
\includegraphics[trim=0cm 0cm 1.5cm 0.5cm, clip=true,width=8.0cm]{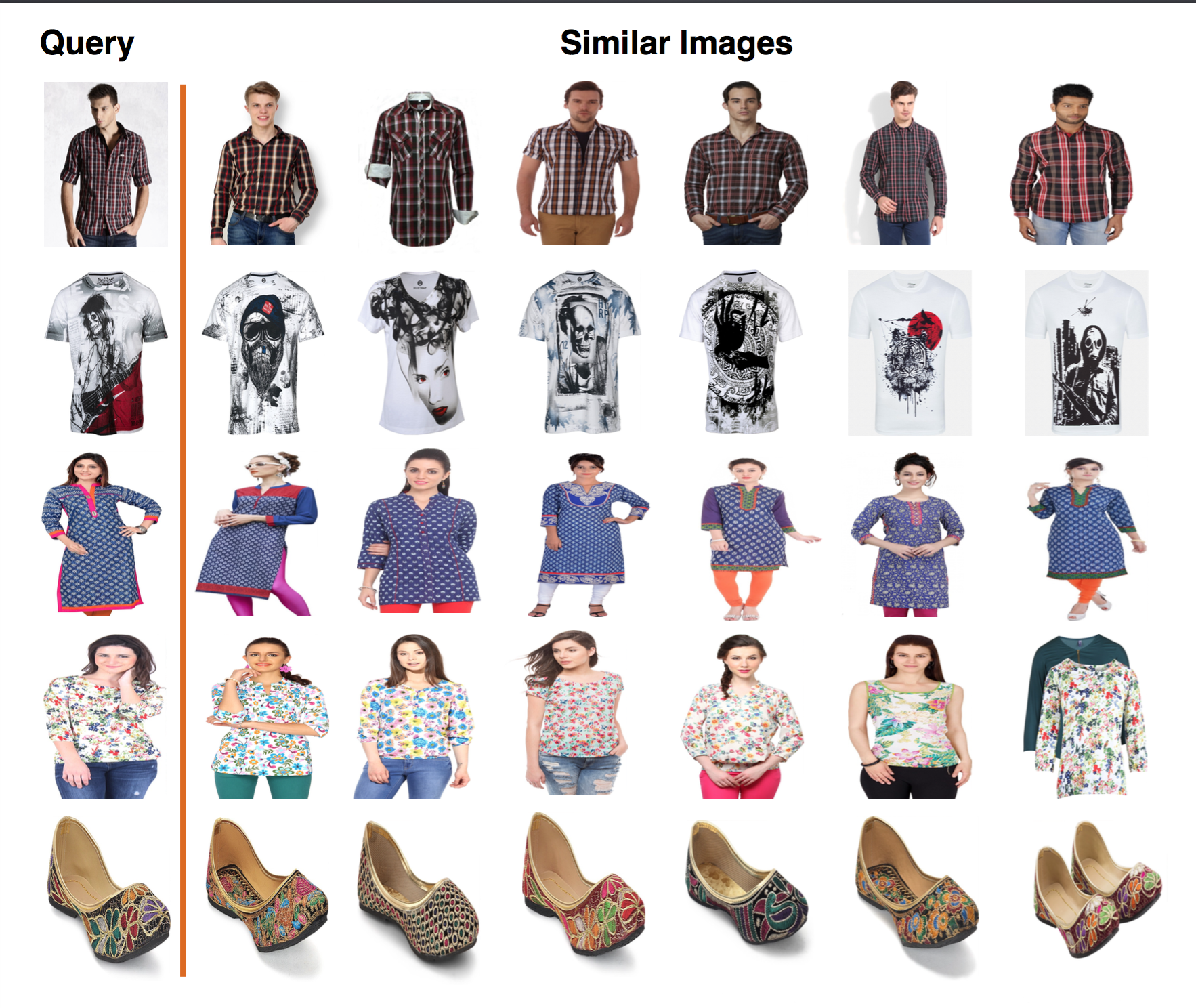}
\vspace{-0.4cm}
\caption{Catalog Image Recommendation Results}
\label{fig:civr_results}
\end{figure}

\begin{table}[]
\centering
\caption{Catalog Image Triplet Accuracies}
\label{CIVRaccuracies}
\begin{tabular}{|p{1.5cm}|p{2cm}|p{2.5cm}|p{1.75cm}|}
\hline
Method  & In-class Triplet Accuracy ($\%$) & Out-of-class Triplet Accuracy ($\%$) & Total Triplet Accuracy ($\%$) \\
\hline
AlexNet  & 71.48  & 91.44  & 84.04 \\
\hline
PatternNet & 67.06  & 87.26 & 79.77 \\
\hline
ColorHist  & 77.45 & 91.32 & 86.18 \\
\hline
\textbf{VisNet} & \textbf{93.30} & \textbf{99.78} & \textbf{97.38} \\
\hline
\end{tabular}
\end{table}

\begin{figure}[t]
\centering
\includegraphics[trim=0cm 0cm 1.5cm 0.5cm, clip=true,width=8.0cm]{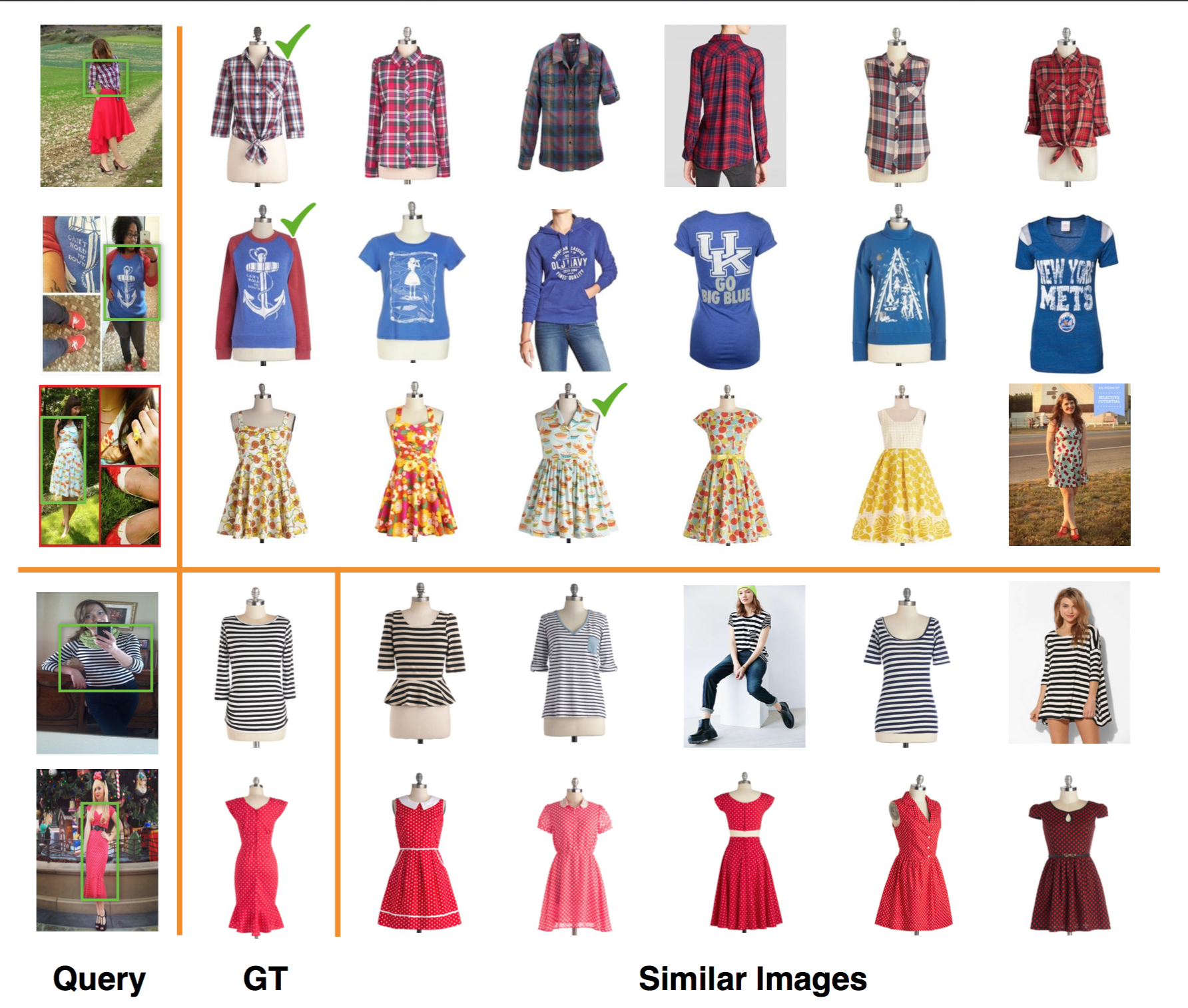}
\caption{Wild Image Recommendation Results from the \textit{Exact Street2Shop} dataset. The top three rows are successful queries, with the correct match marked in green. The bottom two rows are failure queries with ground-truth matches and top retrieval results.}
\label{fig:wil_results}
\vspace{-0.2cm}
\end{figure}

\begin{table*}[]
\centering
\caption{Recall ($\%$) at top-20 on the Exact Street2Shop Dataset}
\label{recallattop20}
\begin{tabular}{|p{1.5cm}|p{1.5cm}|p{2.2cm}|p{2cm}||p{1.8cm}|p{1.5cm}|p{1.5cm}|p{1.3cm}|p{1.3cm}|}
\hline
Category & AlexNet\cite{WhereToBuy} & F.T.~Similarity\cite{WhereToBuy} & R.Contrastive $\&$ Softmax \cite{RobustSiameseVisualSearch} & VisNet-NoShallow & VisNet-AlexNet  & VisNet-S2S & VisNet &  VisNet-\mbox{FRCNN} \\
\hline
Tops & 14.4 & 38.1 & 48.0 & 60.1 & 52.91 & 59.5 & \textbf{62.6} & 55.9 \\
\hline
Dresses & 22.2 & 37.1 & 56.9 & 58.3 & 54.8 & 60.7 & \textbf{61.1} & 55.7 \\
\hline
Outerwear & 9.3 & 21.0 & 20.3 & 40.6 & 34.7 & 43.0 & \textbf{43.1} & 35.9 \\
\hline
Skirts & 11.6 & 54.6 & 50.8  & 66.9 & 66.0 & 70.3 & \textbf{71.8} & 69.8 \\
\hline
Pants & 14.6 & 29.2  & 22.3 & 29.9 & \textbf{31.8} & 30.2 & \textbf{31.8} &  38.5 \\
\hline
Leggings & 14.5  & 22.1 & 15.9 & 30.7 & 21.2 & 30.6 & \textbf{32.4} &  16.9 \\
\hline
\end{tabular}
\end{table*}

\textbf{Flipkart Fashion Dataset}: This is an internally curated dataset that we use for evaluation. As mentioned in section~\ref{ssec:TD}, we programmatically generated candidate training triplets from which the final training triplets were selected through manual vetting. 51K triplets from these were chosen for evaluation (32K out-of-class and 19K in-class). Algorithms were evaluated on what percentage of these triplets were correctly ranked by them (i.e., given a triplet $<q, p, n>$, if $D(q, p) < D(q, n)$, score $1$, else $0$).

The results are shown in Table~\ref{CIVRaccuracies}. We compare the accuracy numbers of our best model, VisNet, with those of the individual BISSs that were used for candidate training triplet generation. The different BISSs that were used are described in Section \ref{ssec:TD}. 

Human evaluation of the results was done before the model was deployed in production. Human operators were asked to rate $5K$ randomly picked results as one of ``Very Bad" , ``Bad", ``Good" and ``Excellent". They rated $97\%$ of the results as ``Excellent". Figure \ref{fig:civr_results} contains some of our results. We also visualise the feature embedding space of our model using the t-SNE \cite{tsne} algorithm. As shown in Figure \ref{fig:tsne}, visually similar items (items with similar colors, patterns and shapes) are clustered together. \\
\textbf{Street2Shop Dataset}: Here we use the Exact Street2Shop dataset from \cite{WhereToBuy}, which contains 20K wild images (street photos) and 400K catalog images(shop photos) of 200K distinct fashion items across 11 categories. It also provides 39K pairs of exact matching items between the street and the shop photos. As in \cite{WhereToBuy}, we use recall-at-k as our evaluation metric - in what percent of the cases the correct catalog item matching the query object in wild image was present in the top-k similar items returned by the algorithm. Our results for six product categories vis-a-vis the results of other state-of-the-art models are shown in Table~\ref{recallattop20} and Figure~\ref{fig:WIVRtopsowdresses}. 

In the tables and graphs, the nomenclature used by us are as follows.
\begin{itemize}
\item \textit{AlexNet:} Pre-trained object recognition AlexNet from caffe model zoo (FC6 output used as embedding).
\item \textit{F.T. Similarity:} Best performing model from \cite{WhereToBuy}. It is a two-layer neural network trained to predict if two features extracted from AlexNet belong to the same item.
\item \textit{R. Contrastive $\&$ Softmax:} Deep Siamese network proposed in \cite{RobustSiameseVisualSearch} which is trained using a modified contrastive loss and a joint softmax loss.
\item \textit{VisNet:}  This is our best performing model. It is a VGG-16 coupled with parallel shallow layers and trained using the hinge loss (see Section \ref{ssec:cnnarch}).
\item \textit{VisNet-NoShallow:} This follows the architecture of VisNet, without the shallow layers.
\item \textit{VisNet-AlexNet:} This follows the architecture of VisNet, with the VGG-16 replaced by the AlexNet. 
\item \textit{VisNet-S2S:} This follows the architecture of VisNet, but is trained only on wild image triplets from the Street2Shop dataset.
\item \textit{VisNet-FRCNN:} This is a two-stage network, the first stage being the object localisation network described in \ref{ssec:localisationVS}, and the second stage being VisNet. These numbers are provided only to enable comparison of the end-to-end system. It should be noted that our object localisation network detects the correct object and its corresponding bounding boxes around 70\% of the time. We report the numbers only for these cases.
\end{itemize}
Our model VisNet outperforms the previous state-of-the-art models on all product categories with an average improvement of around 13.5 \%. From the table, it can be seen that using the VGG-16 architecture as opposed to that of AlexNet significantly improves the performance on wild image retrieval (around 7\% improvement in recall). It can also be seen that the addition of shallow layers plays an important role in improving the recall (around 3\%). Figure \ref{fig:wil_results} presents some example retrieval results. Successful queries (at least one correct match found in the top-20 retrieved results) are shown in the top three rows, and the bottom two rows represent failure cases.

\iftrue
\begin{figure*}[t]
\centering
\includegraphics[trim= 0cm 0cm 0.5cm 0.5cm, clip=true, width=16.0cm]{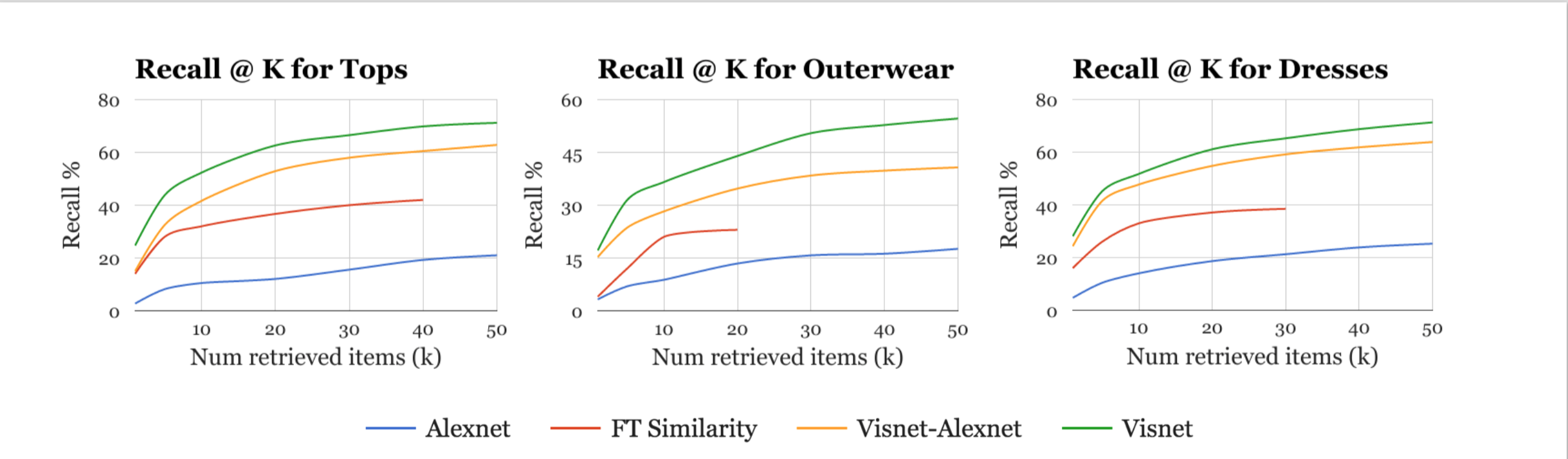}
\caption{Recall at top-k for three product categories on Exact Street2Shop dataset}
\label{fig:WIVRtopsowdresses}
\end{figure*}

\section{Production Pipeline}
\label{sec:Production}

\begin{figure}[t]
\centering
\includegraphics[trim=1.5cm 7cm 0cm 7cm, clip=true, width=8.0cm]{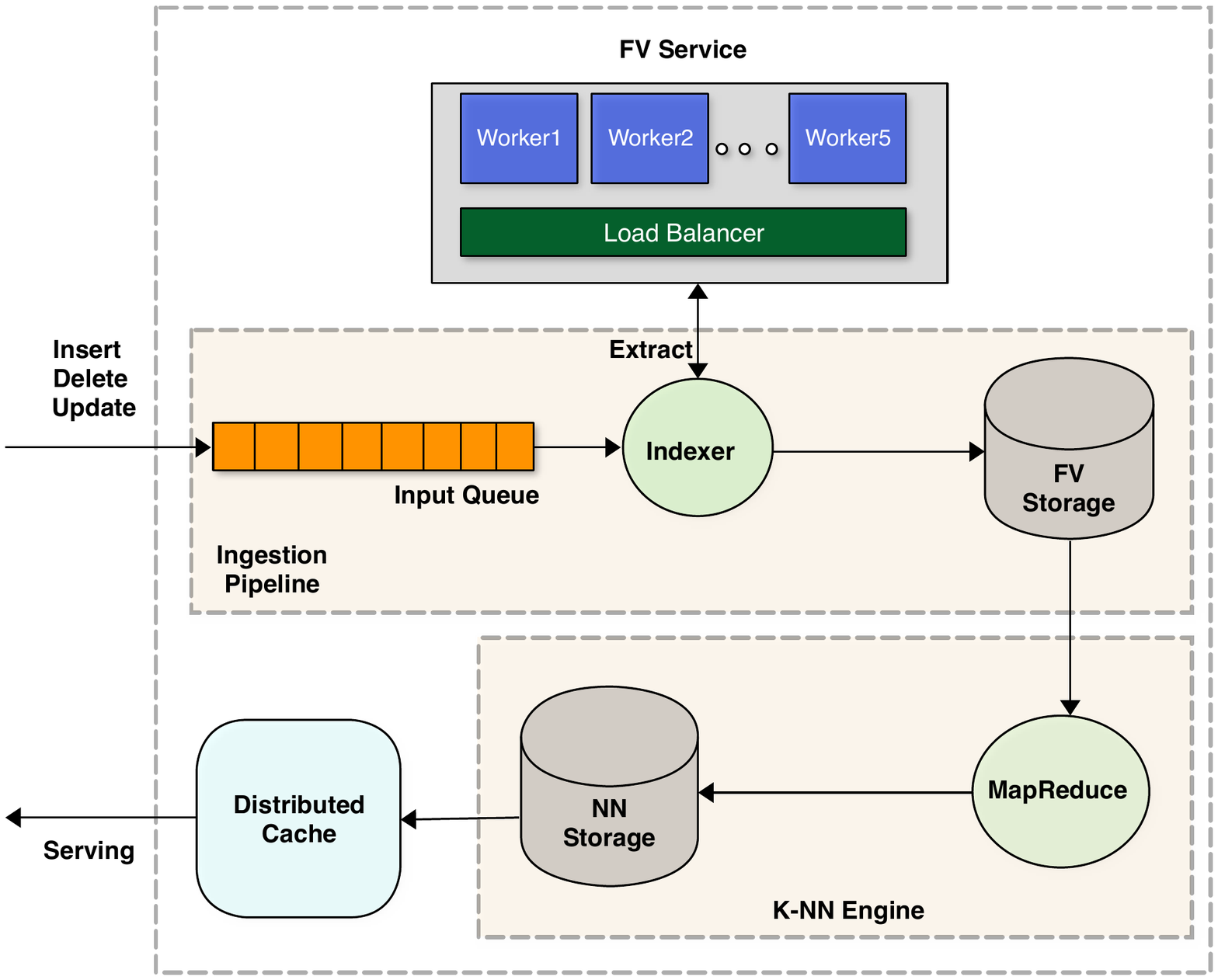}
\caption{Visual Recommendation System}
\label{fig:pipeline}
\end{figure}

We have successfully deployed the Visual Recommendation engine at Flipkart for fashion verticals. In this section, we present the architectural details of our production system, along with the deployment challenges, design decisions and trade-offs. We also share the overall business impact.

\subsection{System Architecture}
The Flipkart catalog comprises of 50 million fashion products with over 100K ingestions (additions, deletions, modifications) per hour. This high rate of ingestion mandates a near real-time system that keeps the recommendation index fresh. We have created a system (see Figure~\ref{fig:pipeline}) that supports a throughput of 2000 queries per second with 100 milliseconds of latency. The index is refreshed every 30 minutes.\\
The main components of the system are 1) a horizontally scalable Feature Vector service,  2) a real-time ingestion pipeline, and 3) a near real-time k-Nearest-Neighbor search. 

\textbf{Feature Vector (FV) Service}: The FV Service extracts the last layer features (embeddings) from our CNN model. Despite the relatively lower performance (see Table \ref{tab:feature_extraction}), we opted for multi-CPU based inferencing to be horizontally scalable in a cost effective manner. An Elastic Load Balancer routes requests to these CPUs optimally. Additional machines are added or removed depending on the load.

\textbf{Ingestion Pipeline}: The Ingestion Pipeline is a real-time system, built using the open-source Apache Storm \cite{Storm} framework. It listens to inserts, deletes and updates from a distributed input queue, extracts feature vectors (for new items) using the FV Service and updates the Feature Vector (FV) Datastore. We use The Hadoop Distributed File System (HDFS \cite{HDFS}) as the choice of our FV Datastore as it is a fault-tolerant, distributed datastore capable of supporting Hadoop Map-Reduce jobs for Nearest-Neighbor search.

\textbf{Nearest Neighbor (NN) Search}: 
The most computationally intensive component of the Visual Recommendation System is the k-Nearest-Neighbour Search across high-dimensional (4096-d) embeddings. Approximate nearest neighbor techniques like Locality Sensitive Hashing (LSH) significantly diminished the quality of our results (by over 10\%), and K-d trees did not produce a performance gain since the vectors are high dimensional and dense. Hence, we resorted to a full nearest neighbor search on the catalog space. As can be seen in Table \ref{tab:knn_computation}, computing the top-k nearest neighbors for 5K items across an index of size 500K takes around 12 hours on a single CPU\footnote{Our CPU experiments were run on 2.3 GHz, 20 core Intel servers.\label{cpu_spec}}. While this can be sped up by multithreading, our scale of operation warrants the k-NN search to be distributed across multiple machines.  

Our k-NN engine is a Map-Reduce system built on top of the open-source Hadoop framework. Euclidean Distance computation between pairs of items happens within each mapper and the ranked list of top-k neighbors is computed in the reducer. To make the k-NN computationally efficient, we make several optimisations. \\
1) \textbf{Incremental MapReduce updates}: Nearest neighbors are computed only for items that have been newly added since the last update. In the same pass, nearest neighbours of existing items are modified wherever necessary. \\
2) \textbf{Reduction of final embedding size}: As can be seen from Table \ref{tab:knn_computation}, k-NN search over 512-d embeddings is significantly faster than over 4096-d embeddings. Hence, we reduced the final embedding size to 512-d by adding an additional linear-embedding layer (4096x512) to VisNet, and retraining. We observed that this reduction in embedding size dropped the quality of results only by a small margin (around 2\%). \\
3) \textbf{Search Space pruning}: We effectively make use of the Metadata associated with catalog items such as vertical (shirt, t-shirt), gender (male, female) to significantly reduce the search space.

While these optimisations make steady-state operation feasible, the initial bootstrapping still takes around 1-week. To speed this up, we developed a GPU-based k-NN algorithm using the CUDA library\footnote{Our GPU experiments were run on nVidia GTX Titan X GPUs\label{gpu_spec}}, that provides a 6x increase in speed over MapReduce (see Table \ref{tab:knn_computation}).

The nearest neighbors computed from the k-NN engine are published to a distributed cache which serves end user traffic. The cache is designed to support a throughput of 2000 QPS under 100 milliseconds of latency. 
\begin{table}[]
\centering
\caption{Feature Extraction Throughput (qps)}
\vspace{-0.2cm}
\label{tab:feature_extraction}
\begin{tabular}{|p{3cm}||p{2cm}|p{2cm}|}
\hline
Network Architecture  & CPU\footref{cpu_spec} & GPU\footref{gpu_spec} \\ \hline\hline
AlexNet + Shallow & 25 & 291   \\ \hline
VGG + Shallow & 5 &  118 \\ \hline
\end{tabular}
\end{table}

\begin{table}[]
\centering
\caption{KNN computation times on 5K x 500K items}
\vspace{-0.2cm}
\label{tab:knn_computation}
\begin{tabular}{|p{2cm}||p{2.5cm}|p{2.5cm}|}
\hline
Method  & 4096-d embeddings & 512-d embeddings \\ \hline\hline
Single-CPU\footref{cpu_spec}  & 12 hours & 9 hours \\ \hline
MapReduce & 2 hours & 30 minutes \\ \hline
Single-GPU\footref{gpu_spec} &  20 minutes & 2 minutes   \\ \hline
\end{tabular}
\end{table}

\subsection{Overall Impact}

\begin{figure}[t]
\centering
\begin{subfigure}[t]{4cm}
\includegraphics[trim=0cm 0cm 0cm 4.6cm,clip=true,width=4.0cm]{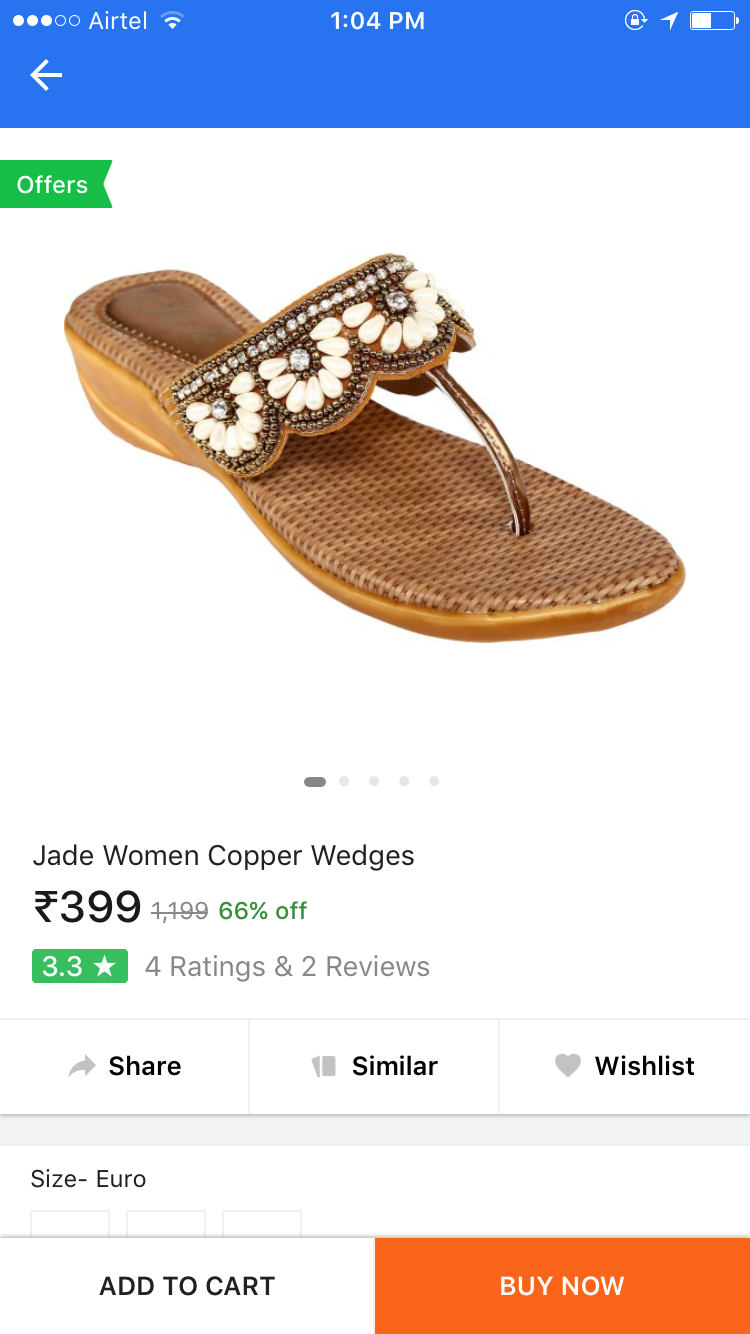}
\caption{Product Page View}
\label{fig:VSAppScreenshot_PPV}
\end{subfigure}
\hfill
\begin{subfigure}[t]{4cm}
\includegraphics[trim=0cm 0cm 0cm 4.6cm,clip=true,width=4.0cm]{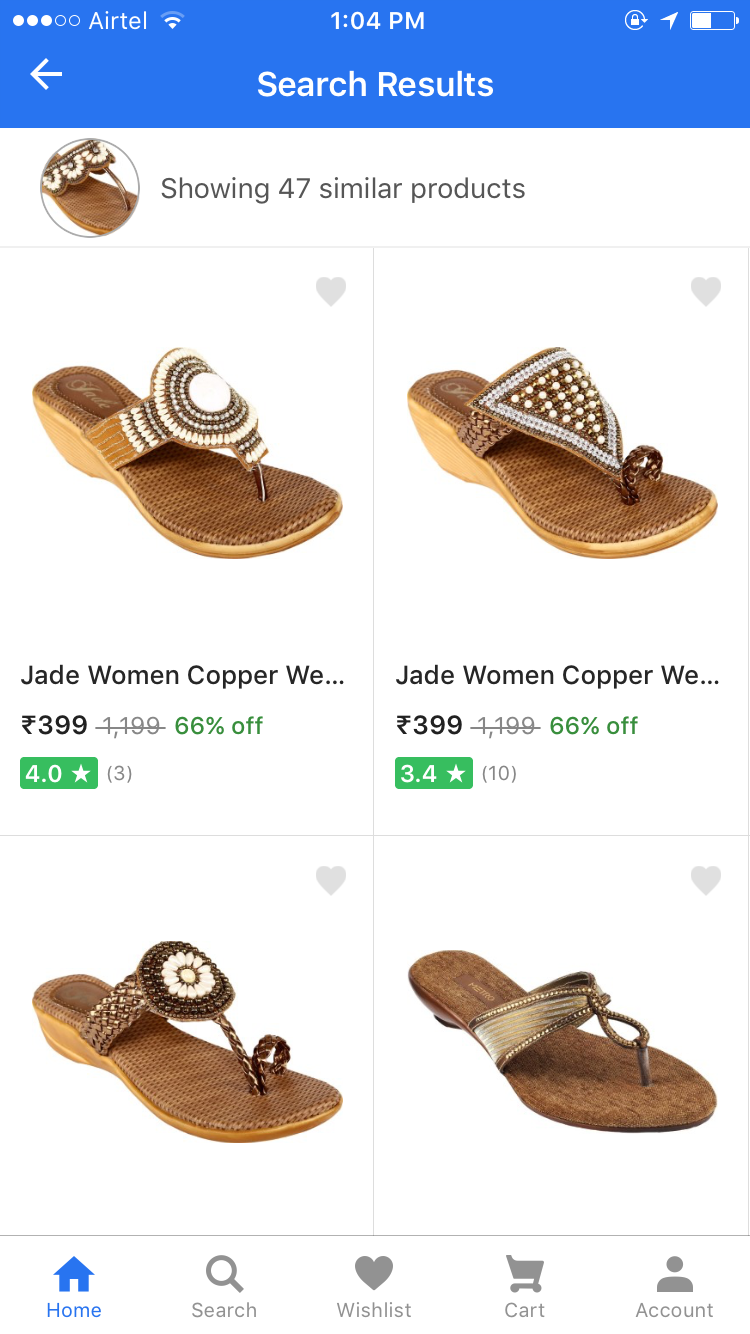}
\caption{Visual Recommendations}
\label{fig:VSAppScreenshot_Similar}
\end{subfigure}
\caption{Visual Recommendations on Flipkart Mobile App}
\label{fig:VSAppScreenshot}
\end{figure}

Our system powers visual recommendations for all fashion products at Flipkart. Figure \ref{fig:VSAppScreenshot} is a screenshot of the visual recommendation module on the Flipkart mobile App. Once a user clicks on the ``similar" button (Figure \ref{fig:VSAppScreenshot_PPV}), the system retrieves a set of recommendations, ranked by visual similarity to the query object (Figure \ref{fig:VSAppScreenshot_Similar}). To measure the performance of the module, we monitor the Conversion Rate (CVR) - percentage of users who went on to add a product to their shopping cart after clicking on the ``similar" button. We observed that Visual Recommendations was one of the best performing recommendation modules with an extremely high CVR of 26\% as opposed to an average CVR of 8-10 \% from the other modules.


In addition to visual recommendations, we have extended the visual similarity engine for two other use cases at Flipkart. Firstly, we have used it to supplement the existing collaborative filtering based recommendation module in order to alleviate the ``cold" start issue. Additionally, we have as used it to eliminate near-duplicate items from our catalog. In a marketplace like Flipkart, multiple sellers upload the same product, either intentionally to increase their visibility, or unintentionally as they are unaware that the same product is already listed by a different seller. This leads to a sub-par user experience. We make use of the similarity scores from the visual recommendation module to identify and remove such duplicates.

\section{Conclusion}
In this paper, we have presented an end-to-end solution for large scale Visual Recommendations and Search in e-commerce. We have shared the architecture and training details of our deep Convolution Neural Network model, which achieves state of the art results for image retrieval on the publicly available Street2Shop dataset. We have also described the challenges involved and various trade-offs made in the process of deploying a large scale Visual Recommendation engine. Our business impact numbers demonstrate that a visual recommendation engine is an important weapon in the arsenal of any e-retailer. 

\begin{figure*}[]
\centering
\includegraphics[trim=0cm 1cm 0cm 1.5cm, clip=true, width=15.0cm]{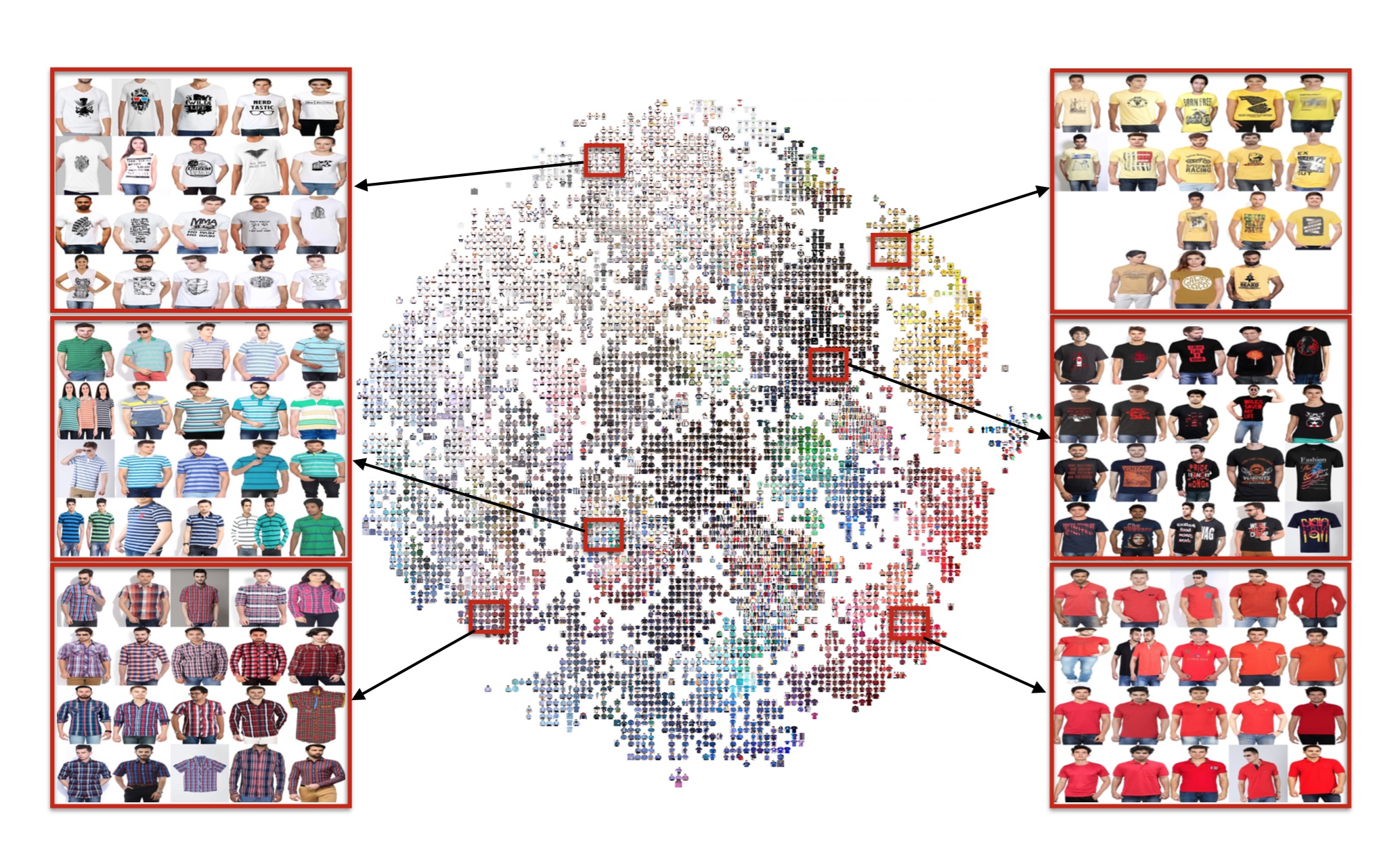}
\caption{t-SNE Visualisation of final layer VisNet embeddings from catalog items}
\label{fig:tsne}
\end{figure*}
\balance
\bibliographystyle{ACM-Reference-Format}
\bibliography{refs} 

\end{document}